\documentclass{article}

\usepackage{arxiv}

\usepackage[utf8]{inputenc} 
\usepackage[T1]{fontenc}    
\usepackage{hyperref}       
\usepackage{url}            
\usepackage{booktabs}       
\usepackage{amsfonts}       
\usepackage{nicefrac}       
\usepackage{microtype}      
\usepackage{lipsum}		
\usepackage{graphicx}
\usepackage{natbib}
\usepackage{doi}
\usepackage{xcolor}

\title{ViT-ReciproCAM: Gradient and Attention-Free Visual Explanations for Vision Transformer}

\author{ 
Seok-Yong~Byun \\
Intel Corp. \\
\texttt{mark.byun@intel.com}  \\
\And
Wonju~Lee \\
Intel Corp.\\
\texttt{wonju.lee@intel.com} \\
}

\renewcommand{\shorttitle}{\textit{arXiv}}

\begin{document}
\maketitle

\begin{abstract}
This paper presents a novel approach to address the challenges of understanding the prediction process and debugging prediction errors in Vision Transformers (ViT), which have demonstrated superior performance in various computer vision tasks such as image classification and object detection. While several visual explainability techniques, such as CAM, Grad-CAM, Score-CAM, and Recipro-CAM, have been extensively researched for Convolutional Neural Networks (CNNs), limited research has been conducted on ViT. Current state-of-the-art solutions for ViT rely on class agnostic Attention-Rollout and Relevance techniques.
In this work, we propose a new gradient-free visual explanation method for ViT, called ViT-ReciproCAM, which does not require attention matrix and gradient information. ViT-ReciproCAM utilizes token masking and generated new layer outputs from the target layer's input to exploit the correlation between activated tokens and network predictions for target classes. Our proposed method outperforms the state-of-the-art Relevance method in the Average Drop-Coherence-Complexity (ADCC) metric by $4.58\%$ to $5.80\%$ and generates more localized saliency maps.
Our experiments demonstrate the effectiveness of ViT-ReciproCAM and showcase its potential for understanding and debugging ViT models. Our proposed method provides an efficient and easy-to-implement alternative for generating visual explanations, without requiring attention and gradient information, which can be beneficial for various applications in the field of computer vision.
\end{abstract}


\section{Introduction}
Recently, vision transformer (ViT) has been proposed by \cite{Alexey2021}, which has gained significant attention in the computer vision community due to its promising representation capabilities. Especially, ViT models such as Deit (\cite{Hugo2020}) have even outperformed traditional convolutional neural networks (CNNs) (\cite{Alex2012,LeCun1989}) in various benchmarks. However, like CNN models, ViT models are susceptible to failure for edge cases or out-of-distribution inputs, which poses a significant problem for critical applications such as medical diagnosis, security, and autonomous driving.

To mitigate this issue, explainable AI (XAI) techniques have been developed for CNN models, such as Class Activation Map (\cite{Zhou2016}), Grad-CAM(\cite{Selvaraju2016}), Recipro-CAM(\cite{Seokyong2022}), Ablation-CAM(\cite{Desai2020}), and Score-CAM(\cite{Wang2020}), which provide high-quality visual explanations using gradient or gradient-free methods. 
However, the research on XAI techniques for ViT is still in its early stages. Only a few approaches have been proposed, such as class agnostic visual activation maps via analyzing attention flow, as suggested by ~\cite{Hila2021}, and the attention-rollout with relevance information to support CAM proposed by ~\cite{Hila2021}. The attention-rollout (\cite{Hila2021}) technique considered skip-connection information and pairwise attention relationship between multi-layer attentions so it improved visualization capability than previous single attention layer analysis methods but it cannot give class specific explainability. ~\cite{Hila2021} improved the visualization quality of attention-rollout and enabled generating class specific activation map with relevance information but it requires gradient information so it cannot be used for supporting explainability in inference time. And, all the existing techniques require the accessibility to the all attention layer's activation in a model and it requires deeper model dependency. 

To overcome the limitations of existing XAI techniques for ViT, we propose an accurate and efficient reciprocal information-based approach. Our method utilizes reciprocal relationship between new spatially masked feature inputs (positional token masking) and network's prediction results. By identifying these relations, we can generate visual explanations that provide insights into the model's decision-making process without using attention layers' information and assuming their internal relationship as previous researches. Our approach not only improves the interpretability of ViT models but also enables users to use XAI result in their inference system without trainable model. We evaluate our method on a range of ViT classification models and demonstrate its effectiveness in generating high-quality visual explanations that aid in understanding the models' behavior and its accuracy in measuring Average Drop-Coherence-Complexity (ADCC) score suggested by ~\cite{Poppi2021}.

The main contributions of this paper include:

\begin{itemize}
\item We present a new, efficient, and gradient-free XAI solution for ViT that utilizes the reciprocal relationship between the spatial masked encoding features and the network's prediction results.
\item The proposed method, ViT-ReciproCAM, achieved a state-of-the-art accuracy on various XAI metrics such as average drop, average increase, and ADCC, while providing approximately $1.5 \times$ faster execution performance than Relevance method.
\end{itemize}

\section{Related Work}
\subsection{Visual Explainability for CNN}
Researches in visual explainability of CNNs have been conducted for a number of years due to their black-box nature, with the goal of revealing the learning mechanism of CNNs. Early research efforts used a deconvolution approach (~\cite{Zeiler2014, Springenberg2015}) to analyze the learned features of each layer and extract important pixels in the input image in a class-agnostic way. This research served as motivation for subsequent white-box XAI researchers. From an explainability perspective, layer-wise activation or input sensitivity measures are insufficient to explain the prediction results of a given model for specific inputs.
The first research that clarified the relationship between input data and CNN output was CAM~\cite{Zhou2016}. CAM produces a feature map that highlights important regions of an image for a target class by multiplying a global average pooling activation vector with a fully connected weight vector specific to the class.
CAM is a useful tool for AI researchers to analyze the neural network architecture and gain an understanding of how the network reacts to specific classes of input data. However, CAM has a limitation in that it requires the presence of a global average or max pooling layer in the architecture. This means that certain neural network architectures may not be compatible with the CAM method, and therefore, alternative explainability methods may need to be considered.

To generalize the CAM technique for all network architectures, gradient-based methods (~\cite{Selvaraju2016, Chattopadhay2018, Fu2020, Omeiza2019}) have been studied by several researchers. These methods rely on the gradient information of the target class confidence output with respect to the activation map of the convolution layer and these operation can be done with training framework's back propagation information.
The gradient-based approaches have emerged as a popular solution, as they can overcome the limitations of CAM and provide high quality visual explainability even though these require a trainable model to utilize gradient information, which restricts their use in post-deployment frameworks like ONNX (~\cite{onnx2019}) or OpenVINO (~\cite{openvino2019}). Moreover, saturation and false confidence issue in gradient-based methods led to the development of gradient-free techniques like Score-CAM (~\cite{Wang2020}), Smooth Score-CAM (~\cite{Wang2021}), Integrated Score-CAM (~\cite{Naidu2020}), Ablation-CAM (~\cite{Desai2020}), and Recipro-CAM (~\cite{Seokyong2022}).

Black-box techniques (~\cite{Petsiuk2018,Kenny2021,Dabkowski2017,Petsiuk2020}) are suggested to remove architectural dependency in white-box methods. These approaches can generate a saliency map without relying on network architectural information or gradient computability. However, they typically require much more time than white-box methods and may exhibit relatively lower accuracy.

\subsection{Visual Explainability for ViT}
The NLP community has conducted extensive research on explainability for the Transformer architecture (~\cite{Ashish2017}). Many studies (~\cite{Jaesong2017, Andy2019, Danish2019, Shikhar2019, Sarah2019, Sarthak2019}) have focused on analyzing attention activation or relations for single attention layers within the self-attention block. A recent work proposed a novel approach to analyzing the relationship between attention layers and consolidating their weights to provide activation scores for each token (~\cite{Samira2020}). However, this method's simple linear relationship assumption between attention layers may not be suitable for abnormally highly activated relations or early attention blocks with noisy information. Additionally, this technique is class-agnostic, and therefore, it cannot provide class-specific explainability.

Building upon the recent proposal of the ViT architecture by ~\cite{Alexey2021}, visual explainability research has commenced, addressing the limitations of the attention-rollout method (~\cite{Samira2020}), which assumes a linear relationship between attention layers and is class-agnostic. The proposed approach employs relevance propagation (~\cite{Gregoire2017}) and gradient information to overcome these shortcomings. With a systematic approach, the authors achieved a significant advance in analyzing attention propagation in the transformer architecture, demonstrating clear differences from the attention-rollout method. Furthermore, they showed that this technique can be utilized for vision transformers with various image inputs in qualitative analysis and appendix. However, it is worth noting that this approach focuses on analyzing attention relations for achieving transformer model explainability and requires a thorough understanding of the given model's architecture and trainable model.    

\begin{figure}
  \begin{center}
    \includegraphics[width=6.0in]{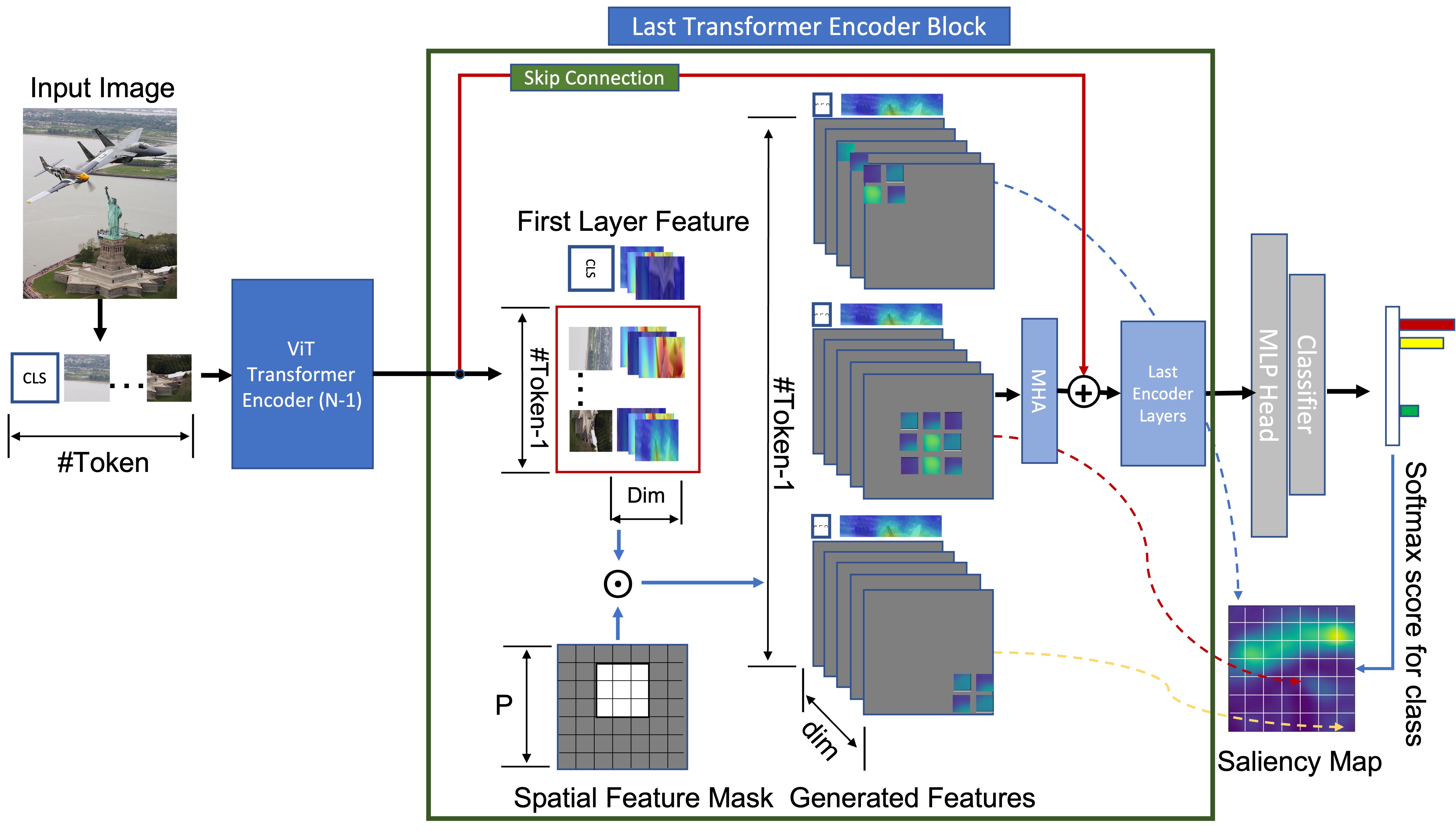}
  \end{center}
  \caption{ViT-ReciproCAM architecture: The first 'LayerNorm' layer output of the last transformer encoder block is used for generating new spatially masked number of patches (\#token-1) feature inputs for using next layers' input using batch operation and normalized prediction score as the target class's saliency map.}
  \label{fig:main}
\end{figure}

XAI solutions must be evaluated with standardized metrics in a unified manner to ensure their effectiveness. However, until the recent proposal of the ADCC metric by ~\cite{Poppi2021}, there was no universally accepted standard metric for XAI evaluation. Previous studies, such as ~\cite{Chattopadhay2018,Petsiuk2018,Fu2020}, had suggested their own metrics such as Avg Drop, Avg Inc, Deletion, and Insertion. Nevertheless, these standalone metrics could lead to incorrect performance evaluations, as demonstrated by ~\cite{Poppi2021} using Fake-CAM example.

ADCC is a comprehensive metric that considers the average drop, coherence, and complexity of the model explanations and provides a single score as their harmonic mean with the following equation:
\begin{equation} \label{eq:adcc}
\textrm{ADCC}(x)= 3\Bigg(\frac{1}{\textrm{Coherency}(x)}+\frac{1}{1-\textrm{Complexity}(x)} +\frac{1}{1-\textrm{AverageDrop}(x)}\Bigg)^{-1}.
\end{equation}

To evaluate the precision of our approach, we will use the ADCC metric, which takes into account multiple aspects of model explanations and provides a more accurate and reliable performance evaluation compared to standalone metrics.

\section{ViT-ReciproCAM}\label{sec:prop}
We propose a novel method, called Vit-ReciproCAM, that efficiently generates saliency maps in ViT without requiring gradient or attention information, as illustrated in Figure~\ref{fig:main}. Specifically, Vit-ReciproCAM extracts a feature map with dimensions $(H \times T \times D)$ from the first layer (LayerNorm) of the last transformer encoder block, where $H$ denotes the number of heads (in the first layer, $H=1$ since all heads are concatenated), $T$ denotes the number of tokens, and $D$ denotes the encoder dimension. From this feature map, Vit-ReciproCAM generates $(T-1)$ spatial masks, each of which corresponds to a new input feature map to be used in the subsequent layer. For each spatial location $(x,y)$ defined by the center of a Gaussian spatial mask, Vit-ReciproCAM measures the prediction scores of a specified class for the corresponding feature token. We note that our explanation ignores the batch dimension for simplicity. The efficacy of Vit-ReciproCAM is evaluated on ImageNet validation dataset and is shown to outperform previous state-of-the-art method.

\subsection{Spatial Token Mask and Feature Generation}
The spatial token mask $M$ has $(N\times T)$ dimensions and where, $N$ is $(T-1)$ and $T$ is '[cls] token' plus number of $Patch(P)^{2}$. This set only $3\times 3$ tokens matched at the spatial positions in patched input image with Gaussian weights, but others set as zero except for the class token for each [0,…,$N$-1] new feature maps as depicted in Figure~\ref{fig:main}.
The new input feature maps $(N\times H\times T\times D)$ can be generated via element-wise multiplication with original feature map $(H\times T\times D)$ and spatial mask $(N\times T)$ as expressed as 

\begin{equation} \label{eq:fm}
\tilde{F}_{k}^{n} = F_{k} \odot M^{n}
\end{equation}
with the Hardamard product $\odot$, where $F_{k}$  is original feature map of the first layer of the last transformer encoding block and ${F}_{k}^{n}$  is new feature maps generated by the operation.

\begin{figure}
  \begin{center}
    \includegraphics[width=5.0in]{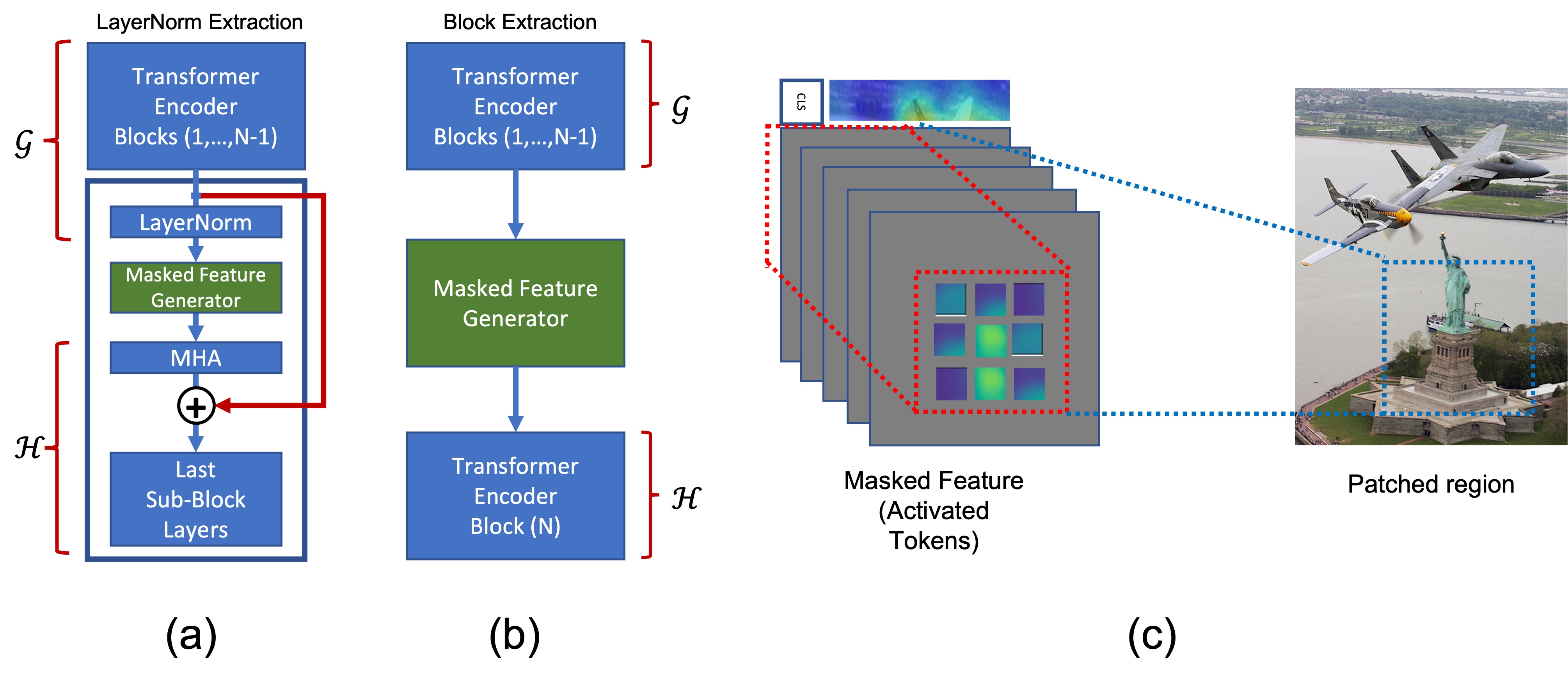}
  \end{center}
  \caption{(a) feature extraction from the first LayerNorm in the last encoder block (default), (b) feature extraction from block output, (c) masked tokens cover the dashed blue box area in input image.}
  \label{fig:fe_mt}
\end{figure}

Figure~\ref{fig:fe_mt}-(c) depicts the process of generating feature maps through a spatial mask. Each feature map comprises $3\times3$ patched regions of the input image with the original features of these regions rescaled by a $3\times3$ Gaussian weight. The resulting masked input feature maps are utilized to compute the saliency score of the center token based on the network's predicted specific class confidence. While the use of the Gaussian kernel for generating the spatial mask is our default approach, it is noteworthy that the mask can also be generated using a single token masking method.


To generate saliency map, ViT-ReciproCAM divides the network into two parts based on the first layer (LayerNorm) of the last transformer encoder block. As depicted in Figure\ref{fig:fe_mt}-(a), the first part is denoted as $\mathcal{G}$, and the subsequent layers are denoted as $\mathcal{H}$. By feeding a batch of $N$ new feature inputs into the last layers ($\mathcal{H}$), we can generate prediction scores for a specified class. These scores indicate the relative importance of the masked token for the class prediction, and the class's saliency map can be calculated using Equation~\ref{eq:Sc}.  

\begin{equation} \label{eq:Sc}
S^c = \textrm{reshape}\left[\frac{\mathbf{Y}_c - \min(\mathbf{Y}_c)}{\max(\mathbf{Y}_c)-\min(\mathbf{Y}_c)}, (P,P) \right],
\end{equation}
where the $N\times 1$ prediction scores $\mathbf{Y}_c=\left[y_{c}^{1}, \dots, y_{c}^{N}\right]^{T}$ for a class $c$ is composed of
\begin{equation} \label{eq:yc}
y_{c}^{n} = \textrm{softmax}\left(\mathcal{H}(\mathcal{G}(I)\odot M^{n})\right)_{c}
\end{equation}
for $n=1,\dots,N$, where $y_{c}$  is a one-dimensional vector of size $N=(P\times P)$ and has $[0, 1]$ scalar value ranges, reshape$[P, Q]$ changes given $P$ dimensional scalar values into $Q$ dimensional scalar values.

\section{Experiments}
In Sections~\ref{sec:quant},~\ref{sec:qual}, and~\ref{sec:perf}, we will provide a quantitative, qualitative, and performance analysis, respectively, to demonstrate the accuracy and effectiveness of ViT-ReciproCAM. 

\subsection{Quantitative Analysis} \label{sec:quant}
To perform quantitative analysis, we utilized the ADCC metric proposed by \cite{Poppi2021}] and conducted experiments under their prescribed conditions. Our evaluation dataset comprised the ILSVRC2012 \cite{Russakovsky2015} validation set, which contains $50,000$ samples. We compared our results to \cite{Hila2021} by evaluating the performance of ViT-ReciproCAM on ViT\_Deit\_Base\_Patch16\_224 and ViT\_Deit\_Small\_Patch16\_224 (\cite{Hugo2020}). Prior to feeding the images into the model, we resized them to $256 \times 256$ and center-cropped them to $224 \times 224$. Furthermore, we normalized the images using a mean of $[0.485, 0.456, 0.406]$ and a standard deviation of $[0.229, 0.224, 0.225]$. The outcomes of our analysis are presented in Table~\ref{tb:perf}. ViT-ReciproCAM achieved state-of-the-art ADCC score on two ViT architectures.

\begin{table}
\caption{Comparison of ViT-ReciproCAMs and Relevance using ADCC metric on two ViT architectures.}
\scriptsize \centering
\begin{tabular}{lccccc|ccccc} \hline
& \multicolumn{5}{c|}{Deit\_Base\_Patch16\_224} & \multicolumn{5}{c}{Deit\_Small\_Patch16\_224} \\ \hline
Method & \begin{tabular}[c]{@{}c@{}} \textbf{Drop} \\ $(\downarrow)$ \end{tabular} & \begin{tabular}[c]{@{}c@{}}Inc \\ $(\uparrow)$ \end{tabular} & \begin{tabular}[c]{@{}c@{}} \textbf{Coher} \\ $(\uparrow)$ \end{tabular} & \begin{tabular}[c]{@{}c@{}} \textbf{Compl} \\ $(\downarrow)$ \end{tabular} & \begin{tabular}[c]{@{}c@{}} \textbf{ADCC} \\ $(\uparrow)$ \end{tabular} & \begin{tabular}[c]{@{}c@{}} \textbf{Drop} \\ $(\downarrow)$ \end{tabular} & \begin{tabular}[c]{@{}c@{}}Inc \\ $(\uparrow)$ \end{tabular} & \begin{tabular}[c]{@{}c@{}} \textbf{Coher} \\ $(\uparrow)$ \end{tabular} & \begin{tabular}[c]{@{}c@{}} \textbf{Compl} \\ $(\downarrow)$ \end{tabular} & \begin{tabular}[c]{@{}c@{}} \textbf{ADCC} \\ $(\uparrow)$ \end{tabular} \\ \hline
Relevance[\cite{Hila2021}] & 55.31 & 9.30 & 79.41 & 13.17 & 64.53 & 55.85 & 10.00 & 80.67 & 12.53 & 64.54 \\
ViT-ReciproCAM & 33.57 & 9.78 & 70.22 & 53.54 & 59.03 & 31.42 & 11.87 & 67.62 & 54.21 & 58.58 \\
\textbf{ViT-ReciproCAM $[3\times 3]$} & 25.82 & 14.12 & 88.61 & 46.35 & \textbf{69.11} & 27.69 & 17.12 & 84.58 & 41.16 & \textbf{70.34} \\ \hline
\end{tabular} \label{tb:perf}
\end{table}

\subsection{Qualitative Analysis}\label{sec:qual}
In this study, we employed the ImageNet pretrained ViT\_Deit\_Base\_Patch16\_224 model (\cite{Hugo2020}) provided by Facebook Research. We used the shared source provided by ~\cite{Hila2021} to evaluate the Relevance method's saliency map. To ensure consistency in the quantitative analysis, we applied identical pre-processing and normalization techniques to the input images. The input images were categorized into three groups for systematic analysis: single-object images, multiple identical objects in the input images, and input images with multiple classes.

\begin{figure}
\begin{center}
\includegraphics[width=6.5in]{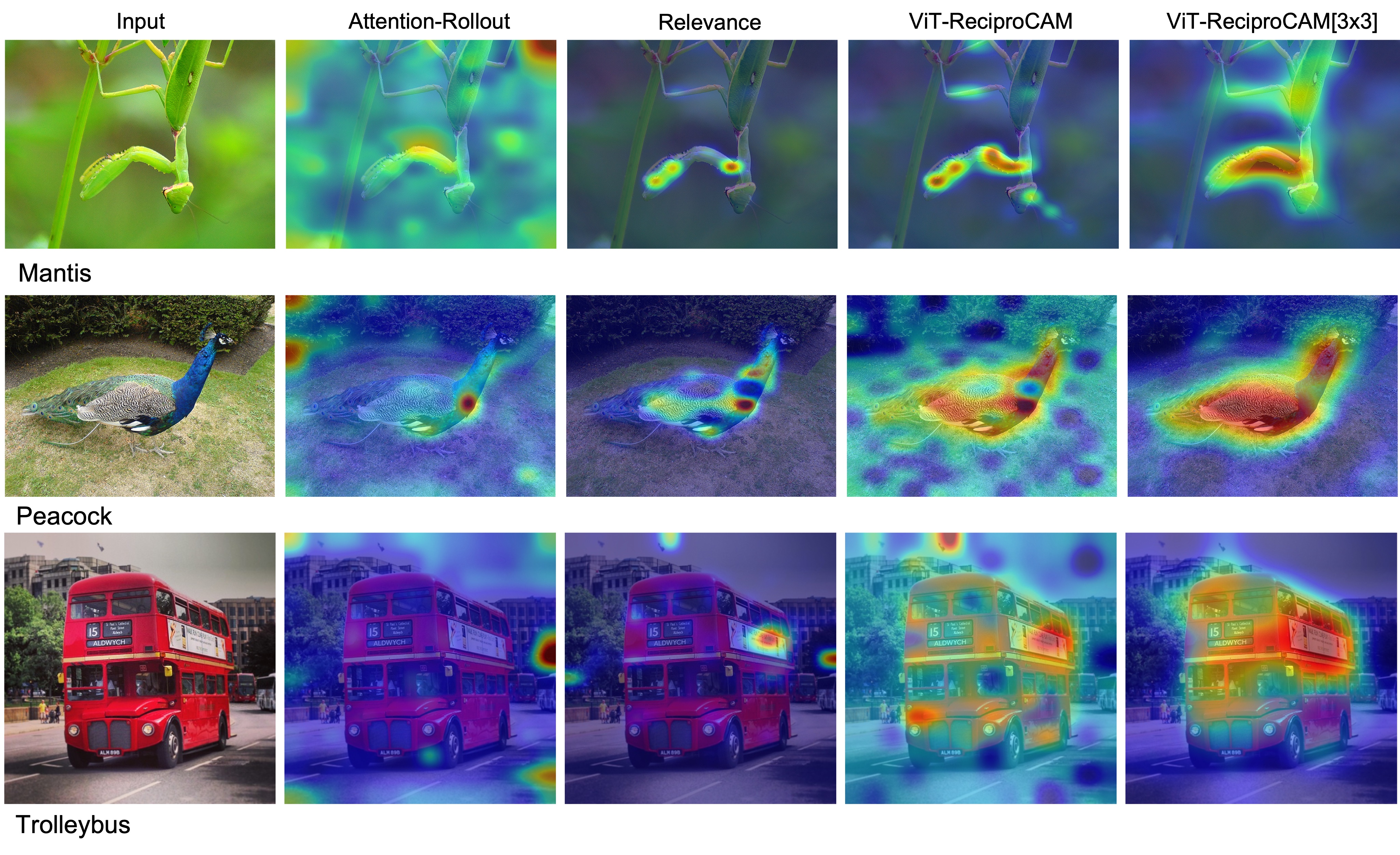}
\end{center}
\caption{Single-object results: Mantis and Peacock inputs process with Attention-Rollout, Relevance, ViT-ReciproCAM, and ViT-ReciproCAM $[3\times 3]$ to generate saliency maps.}
\label{fig:ex1}
\end{figure}

In Figure~\ref{fig:ex1}, we present the results of the initial case comparison. The Attention-Rollout method (\cite{Samira2020}) produces a class-agnostic output with a high level of noise in the saliency map. In contrast, the Relevance (\cite{Hila2021}), ViT-ReciproCAM, and ViT-ReciproCAM $[3\times 3]$ methods exhibit precise activation of the 'mantis', 'peacock', and 'trolleybus' classes. However, the overall quality of Relevance's saliency map is inferior to that of ViT-ReciproCAM and ViT-ReciproCAM $[3\times 3]$ approaches, although ViT-ReciproCAM shows some activated background regions for 'peacock' and 'trolleybus'.

\begin{figure}
\begin{center}
\includegraphics[width=6.5in]{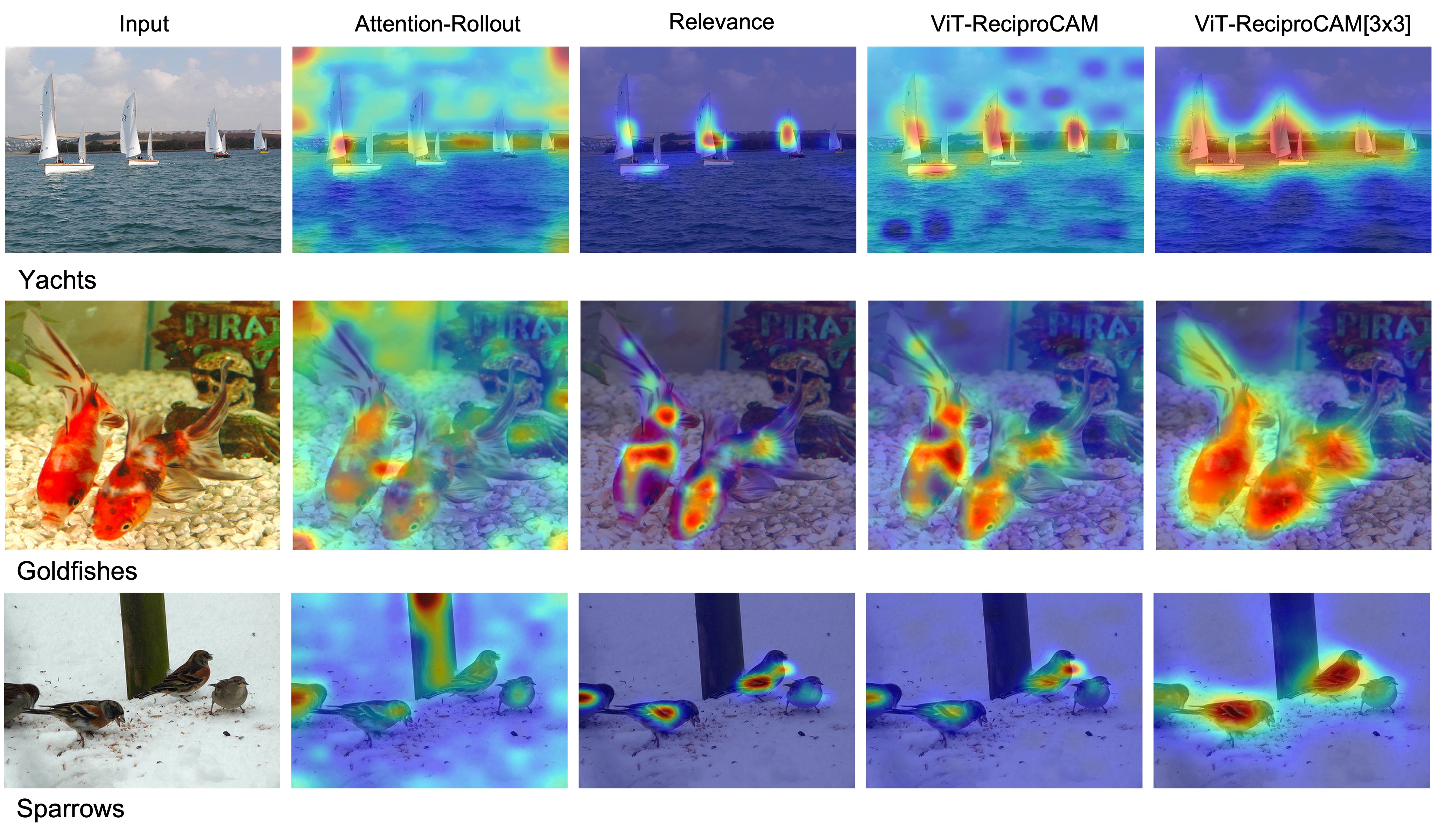}
\end{center}
\caption{Same multiple-objects results: Yachts, Goldfishs, and Sparrows inputs process with Attention-Rollout, Relevance, ViT-ReciproCAM, and ViT-ReciproCAM $[3\times 3]$ to generate saliency maps.}
\label{fig:ex2}
\end{figure}

The second experiment aims to evaluate the localization performance of the proposed methods. The results are presented in Figure~\ref{fig:ex2}. Compared to the Relevance approach, which activates only limited key features of each object through a narrow range of the saliency map, the ViT-ReciproCAM family exhibits wider coverage of the class object regions. Notably, the ViT-ReciproCAM $[3\times 3]$ method yields highly localized saliency maps for the target class objects.

\begin{figure}
\begin{center}
\includegraphics[width=6.5in]{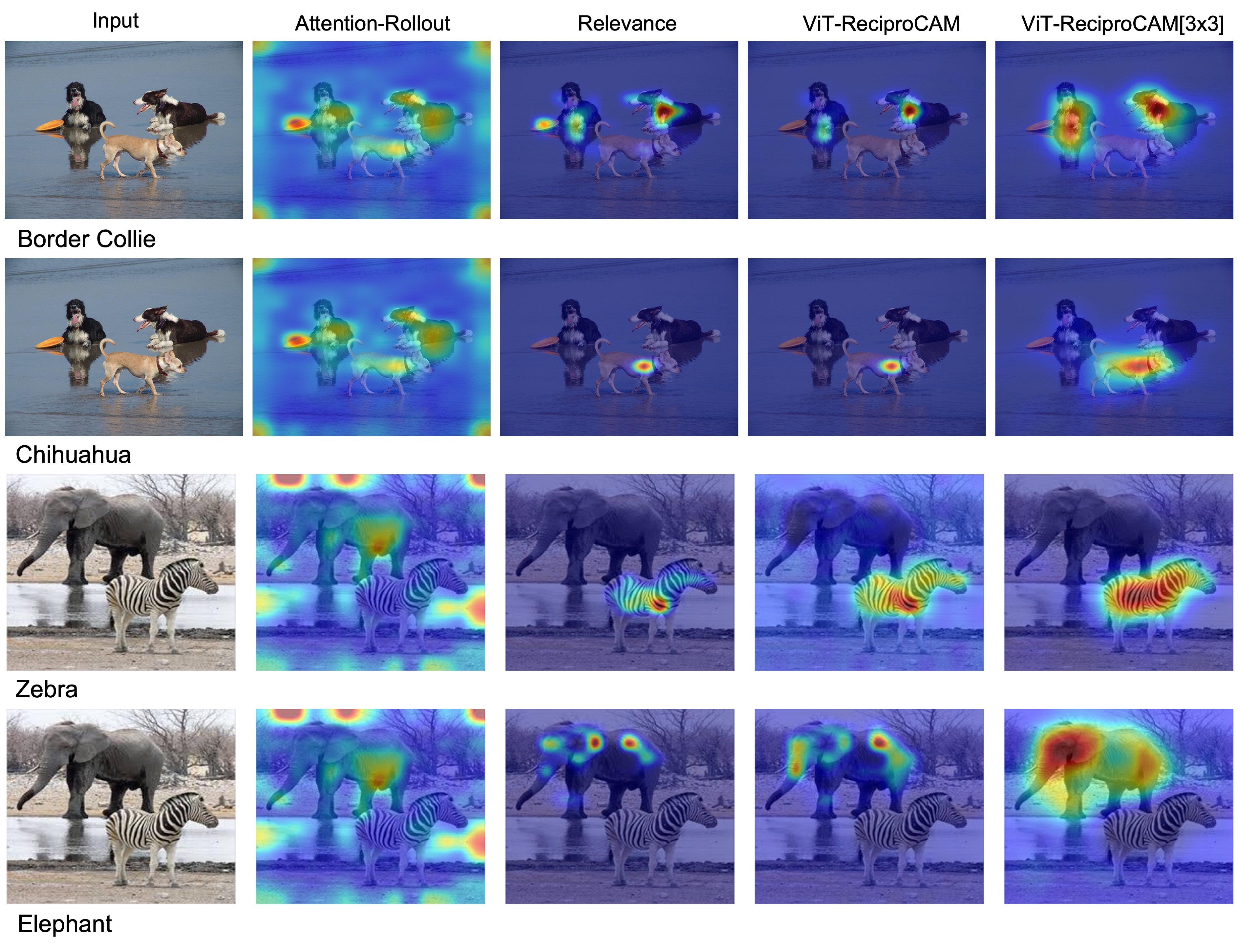}
\end{center}
\caption{Multiple-class results: Border Collie/Chihuahua and Zebra/Elephant inputs process with Attention-Rollout, Relevance, ViT-ReciproCAM, and ViT-ReciproCAM $[3\times 3]$ to generate each target class's saliency maps.}
\label{fig:ex3}
\end{figure}

In third experiment, we evaluate the resolution capability of different XAI methods across multiple object classes in an image. To achieve this, we deactivate objects from different classes within an image, allowing the resulting saliency map to reflect class-specific results. The comparison of different methods is presented in Figure~\ref{fig:ex3}. The first row of Figure~\ref{fig:ex3} shows the results obtained for an image of a 'border collie' containing a 'chihuahua' object, where the network predicted 'border collie' with the highest probability. While Relevance, ViT-ReciproCAM, and ViT-ReciproCAM $[3\times 3]$ successfully separate the different class object(s) from the target class object(s), Attention-Rollout fails to activate the target object(s). In the second row, we generate a saliency map for the 'chihuahua' class to evaluate the quality of the saliency maps generated for other classes in a multiple classes input case. The third and fourth rows show another multiple classes case for 'Elephant' and 'Zebra' using ImageNet's image normalization method described in Section~\ref{sec:quant}. All three methods except for Attention-Rollout successfully localize each class. However, the quality of the saliency map generated by the ViT-ReciproCAM family is superior to Relevance.

\subsection{Performance Analysis}\label{sec:perf}
We conducted performance measurement experiments on Relevance, ViT-ReciproCAM, and ViT-ReciproCAM $[3\times 3]$, using the Deit\-base\-patch16\-224 ImageNet pre-trained model. The test dataset consisted of $1,000$ randomly selected images from the ILSVRC2012 validation dataset, and the image preprocessing and crop size were consistent with the quantitative analysis process. The experimental setup included a single Nvidia Geforce RTX 3090 device and an i9-11900 Intel CPU. As indicated in Table~\ref{tb:speed}, Relevance exhibited a performance slowdown of approximately 1.5x compared to ViT-ReciproCAM Family.   

\begin{table}
\caption{A comparative analysis of the execution times of three distinct methodologies. The measurement of execution time was conducted using $1,000$ inputs, and the resultant average time was calculated.}
\footnotesize \centering
\begin{tabular}{llll} \hline
& \begin{tabular}[c]{@{}l@{}}Execution\\ Time (ms) \end{tabular} & FPS & Ratio   \\ \hline
ViT-ReciproCAM & 86.4 & 11.57 & - \\
\textbf{ViT-ReciproCAM $[3\times 3]$} & 88.1 & 11.35 & 1.02$\times$ \\ 
Relevance   & 132.2 & 7.56 & 1.50$\times$  \\ \hline
\end{tabular} \label{tb:speed}
\end{table}

\section{Ablation Study}
In this section, we present two experiments aimed at assessing the impact of the class token ([CLS] token) on spatially masked feature maps, and the dependency of feature extraction position in transformer encoder blocks. We evaluated the experiments by measuring ADCC scores for two cases: without the class token (wo\_CLS\_token), 
by extracting the feature map from output of the $(N-1)$ block (block[-2]) using ViT\_Deit\_Base\_Patch16\_224, ViT\_Deit\_Small\_Patch16\_224, and ViT\_Deit\_Tiny\_Patch16\_224 ViT architectures. The experimental conditions were kept the same as those used in Section \ref{sec:quant}.

\subsection{Class Token Effect}\label{sec:cls_mask}
\begin{figure}
\begin{center}
\includegraphics[width=6.5in]{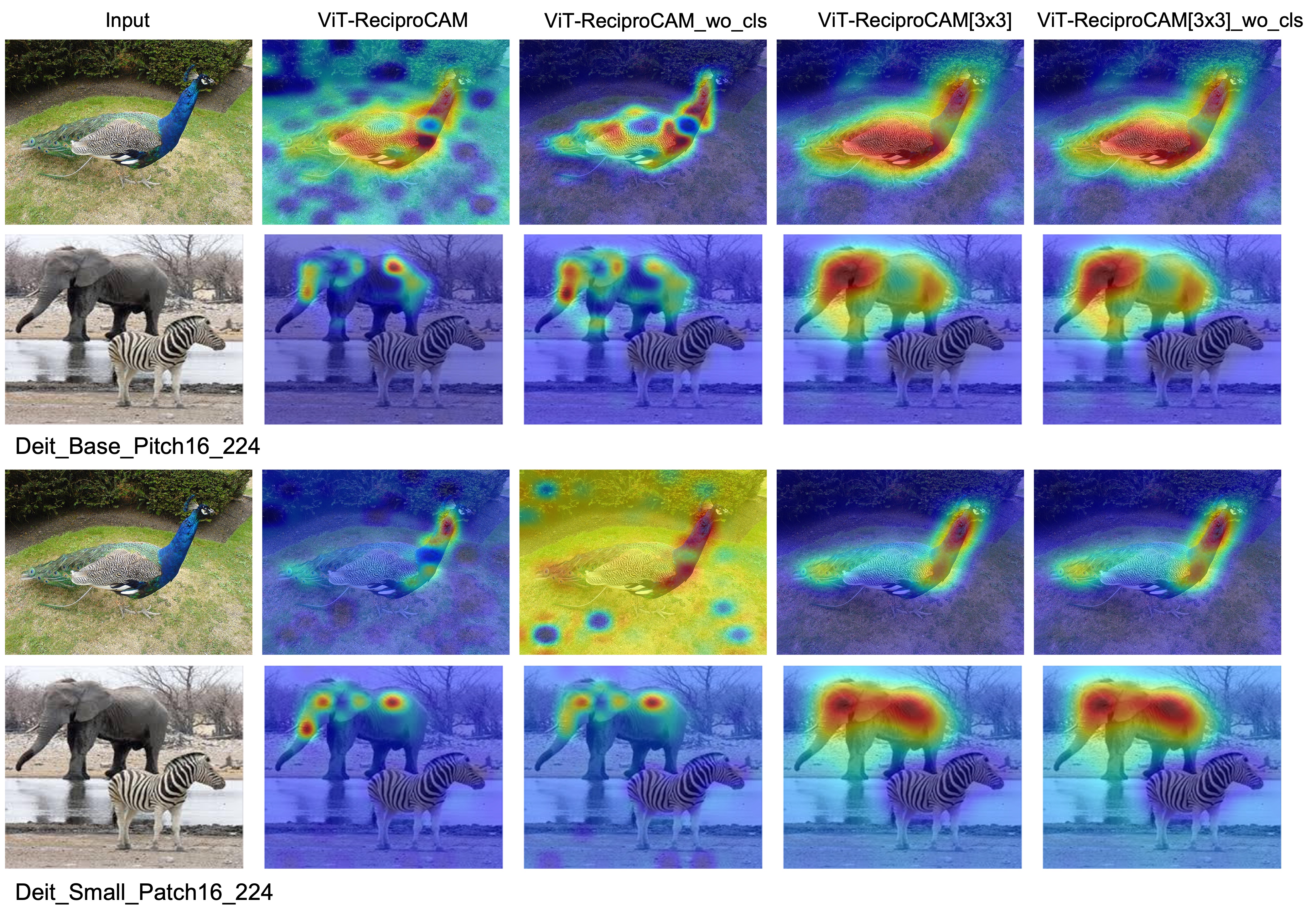}
\end{center}
\vspace{-3mm}
\caption{The impact of class token information on different capacity of ViT architectures.}
\label{fig:abl_cls}
\vspace{-2mm}
\end{figure}

In order to investigate the impact of class token information, we modified the default spatial mask generation logic to zero out the [cls] token feature values, thus reducing the effect of the [cls] token in skip-connected feature maps after the multi-head attention (MHA) layer.
We compared the results of the modified model to the default model with the class token included, and summarized the findings in Table~\ref{tb:abl_cls}.
Eliminating class token information from spatially masked new feature maps produced different effects depending on the capacity of ViT architectures and the kernel method used (Gaussian [$3\times3$] and Dirac Delta).
In the Base model, eliminating the [cls] token information resulted in higher ADCC scores compared to the default cases, but the score gap between the models was smaller for the Gaussian kernel method than the Dirac Delta kernel method. However, in the other two low-capacity architectures, the ADCC scores for the eliminated cases were lower than the default cases. The score gap tendencies between the two cases in the two kernel methods were consistent with the high-capacity architecture. Therefore, the [cls] token information can serve as a helper to maintain the stability of ViT-ReciproCAM in an architecture-agnostic manner. 
The qualitative comparison is depicted in Figure~\ref{fig:abl_cls}.     

\begin{table}
\caption{Ablation study results: effect of class token}
\footnotesize \centering
\begin{tabular}{lrrrrrr} \hline
& Base & (Diff) & Small & (Diff) & Tiny & (Diff)   \\ \hline
\textbf{ViT-ReciproCAM} & 59.03 & - & \textbf{58.58} & - & \textbf{60.00} & - \\
ViT-ReciproCAM\_wo\_cls\_token & \textbf{67.05} & \textcolor{teal}{8.02} & 48.03 & \textcolor{purple}{-10.55} & 56.82 & \textcolor{purple}{-3.18} \\ \hline
\textbf{ViT-ReciproCAM $[3\times 3]$} & 69.11 & - & \textbf{70.34} & - & \textbf{70.37} & - \\
ViT-ReciproCAM $[3\times 3]$\_wo\_cls\_token   & \textbf{69.58} & \textcolor{teal}{0.47} & 69.51 & \textcolor{purple}{-0.83} & 69.78 & \textcolor{purple}{-0.59} \\ \hline
\end{tabular} \label{tb:abl_cls}
\end{table}

\subsection{Effect of Feature Extraction Position}\label{sec:fe_position}
In our default approach, we extract feature information from the first 'LayerNorm' layer of the last transformer encoder block in the network. This is because utilizing the unmodified skip connected feature information from the previous encoder block can provide more accurate explainability. This method is compared with a block feature extraction method illustrated in Figure~\ref{fig:fe_mt}-(b) to show superiority of it and the result is summarized in Table~\ref{tb:ablation_block}.

\begin{table}
\caption{Ablation study results: effect of feature extraction position}
\footnotesize \centering
\begin{tabular}{lrrrrrr} \hline
& Base & (Diff) & Small & (Diff) & Tiny & (Diff)   \\ \hline
\textbf{ViT-ReciproCAM} & \textbf{59.03} & - & \textbf{58.58} & - & \textbf{60.00} & - \\
ViT-ReciproCAM\_block & 52.92 & \textcolor{purple}{-6.11} & 58.46 & \textcolor{purple}{-0.12} & 59.77 & \textcolor{purple}{-0.23} \\ \hline
\textbf{ViT-ReciproCAM $[3\times 3]$} & \textbf{69.11} & - & \textbf{70.34} & - & \textbf{70.37} & - \\
ViT-ReciproCAM $[3\times 3]$\_block   & 67.17 & \textcolor{purple}{-1.94} & 66.56 & \textcolor{purple}{-3.78} & 65.92 & \textcolor{purple}{-4.45} \\ \hline
\end{tabular} \label{tb:ablation_block}
\end{table}


\section{Conclusion}
In this paper, we proposed a novel approach called Vit-ReciproCAM to generate visual explainability for vision transformers (ViT) without requiring gradient and attention information. Vit-ReciproCAM masks the feature map of the first layer of the last transformer encoder block to leverage the correlation between the spatially masked feature maps and the network outputs. Our experimental results demonstrate that Vit-ReciproCAM outperforms other saliency map generation methods in terms of both performance and execution time, achieving state-of-the-art results $(69.11/70.34)$ on the ADCC metric. Remarkably, Vit-ReciproCAM's execution time is approximately 1.5$\times$ faster than that of the attention and gradient-based Relevance method. The proposed approach could potentially aid in improving model interpretability and assist practitioners in understanding how vision transformers make decisions. 

\bibliographystyle{unsrtnat}
\bibliography{references}  

\end{document}